\pdfoutput=1
\documentclass[11pt]{article}

\usepackage[]{EACL2023}

\usepackage{times}
\usepackage{latexsym}
\usepackage{amsmath}
\usepackage{breqn}
\usepackage{amsfonts}
\usepackage{multirow}
\usepackage{booktabs}
\usepackage{algorithm}
\usepackage{subcaption}
\usepackage{arydshln}

\usepackage{colortbl}
\newcommand{\CC}[1]{\cellcolor{blue!#1}}
\newcommand{\CCR}[1]{\cellcolor{red!#1}}
\newcommand{\CCG}[1]{\cellcolor{green!#1}}

\usepackage{ragged2e}
\definecolor{esc}{RGB}{0, 153, 51}

\usepackage[T1]{fontenc}
\usepackage[utf8]{inputenc}

\usepackage{microtype}

\usepackage{inconsolata}

\title{Reward Engineering for Generating Semi-structured Explanation}

\author{
  Jiuzhou Han$^{{\natural} }$\ \ \ \ \ 
  Wray Buntine$^{{\flat}}$\ \ \ \ \ 
  Ehsan Shareghi$^{{\natural} }$\\
  $^{{\natural} }$~Department of Data Science \& AI, Monash University \\
  $^{{\flat}}$~College of Engineering and Computer Science, VinUniversity\\
  {jiuzhou.han@monash.edu}\ \ \ \ \ \ {wray.b@vinuni.edu.vn} \\{ehsan.shareghi@monash.edu}
}

\begin{document}
\maketitle
\begin{abstract}
%\begin{itemize}
%    \item Importance of explainable AI/Predictive modelling (may be see how the explagraph paper builds its narrative)
%    \item Then transition into (semi-)structured explanation as a pivotal framing of explainibity (i.e., why should the reader care about structured explainability).
%\end{itemize}
%Large Language Models (LLMs) have shown great capability in complex reasoning tasks across a wide range of fields. \ehsan{Despite their success at producing a correct output, evaluating true capabilities of model for arriving at a correct output, requires an explainable  mechanism, often presented in a (semi-)structured form.} %Investigating the explainability of LLMs is useful also significant. 

%generation (SEG) aims to represent the internal reasoning process behind a model's prediction in a semi-structured format. This kind of structured explanation is intended to

Semi-structured explanation depicts the implicit process of a reasoner with an explicit representation. This explanation highlights how available information in a specific query is utilised and supplemented with information a reasoner produces from its internal weights towards generating an answer. Despite the recent improvements in generative capabilities of language models, producing structured explanations to verify a model's true reasoning capabilities remains a challenge. This issue is particularly pronounced for not-so-large LMs~(e.g., FLAN-T5-XXL).
%, as the reasoner is expected to couple a sequential answer with a structured explanation which embodies both the \emph{correct presentation} and the \emph{correct reasoning process}. 
%
%Reliance on models' internal implicit knowledge to generate external information has potential to introduce various forms of misalignment with the task goal. 
%
In this work, we first underscore the limitations of supervised fine-tuning~(SFT) in tackling this challenge, and then introduce a carefully crafted reward engineering method in reinforcement learning~(RL) to better address this problem. We investigate multiple reward aggregation methods and provide a detailed discussion which sheds light on the promising potential of RL for future research. Our proposed method on two semi-structured explanation generation benchmarks~(ExplaGraph and COPA-SSE) achieves new state-of-the-art results.~\footnote{Our code is available at \url{https://github.com/Jiuzhouh/Reward-Engineering-for-Generating-SEG}.}

%... this task is difficult because some sub-graph information should be inferred by the model, which is not provided in the input. In this paper, we propose to use the designed reward engineering method in reinforcement learning to better address this problem. We investigate multiple reward aggregation methods and provide insightful analysis on this task. We conduct experiments on two semi-structured explanations generation datasets and our proposed method achieves a new state-of-the-art result. 
\end{abstract}

% (1) Despite the strands of work on guiding LLMs to generate reasoining through a trajectory of reasoning steps (cite CoT/ToT/etc), these trajectories are hard to evaluate. As an orthoghonal approach to explainable reasoning trajectories, structured explanation attempts to produce a reasoning format which justifies model's prediction.   
% Is CoT same as explanation graph? Is explagraph more difficult than CoT? What advantage explagraph has over CoT?
\section{Introduction}
% \begin{itemize}
%     \item motivate the importance of SEG and its difficulty 1p
%     \item existing solutions and what is missing: no single model solution, LMs or LLMs even strugle at generating structured output alone (PIVE). (1-2p) most important part of the paper.
%     \item we present ... (1P)
%     \item findings + punchy closing statement (1P)
% \end{itemize}

\begin{figure}[t]
    \centering
    \includegraphics[scale=0.7,trim={5.8cm 1cm 5.5cm 0cm},clip]{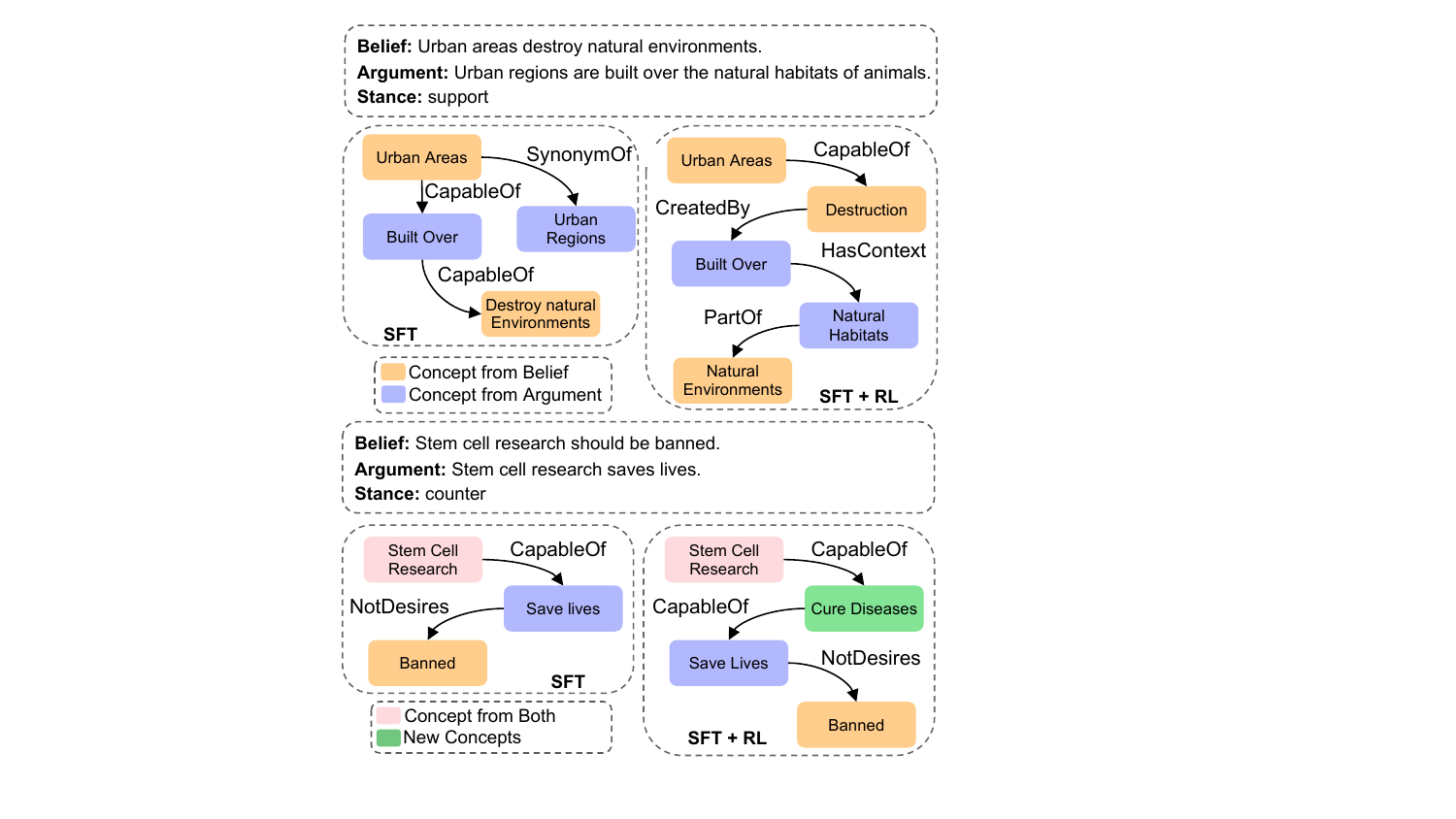}
    \caption{Given the belief and argument, the task is to predict the stance (support/counter) and generate an explanation graph representing the reasoning process. The explanation graph under SFT+RL is more expressive.}
    \vspace{-3mm}
    \label{fig:seg}
\end{figure}

Language models have shown great capability in complex reasoning tasks~\citep{DBLP:journals/corr/abs-2307-09288,Bubeck2023SparksOA,DBLP:journals/corr/abs-2302-13971,DBLP:journals/corr/abs-2210-11416,brown2020language,DBLP:conf/emnlp/Yang0ZBCSM18, DBLP:conf/emnlp/LinCCR19}. Despite their proficiency in generating accurate results, a comprehensive assessment of the models' true capabilities in reaching the correct output necessitates an explainable mechanism. 
%In another word, it's vital to ascertain whether the model is arriving at the correct answer for the right reasons. %Among many approaches in the realm of explainability~\citep{DBLP:journals/csur/MadsenRC23, doshi2017towards}, 
In this spirit, generating structured explanations~\citep{DBLP:conf/emnlp/SahaYBB21,DBLP:conf/lrec/BrassardHKI22} is a viable approach as they explicitly representing the relationships between facts employed during reasoning, and are amenable to evaluation.  
Unstructured natural language explanations lack these aspects.
Figure~\ref{fig:seg} illustrates two examples of stance detection task, where the structured outputs are intended to explain the stance.

%Generation of a reasoning path in large language models (LLMs) like the GPT family~\citep{DBLP:journals/corr/abs-2303-08774}, have been mostly orchestrated by Chain-of-Thought~(CoT)~\citep{DBLP:conf/nips/KojimaGRMI22,DBLP:conf/nips/Wei0SBIXCLZ22}, or more  recently Tree-of-thought~\citep{DBLP:journals/corr/abs-2305-10601} and Graph-of-Thought~\citep{DBLP:journals/corr/abs-2305-16582} approaches. These approaches enable LLMs to generate an internal reasoning process before producing a response. While this has a proven impact on improving models' predictive capabilities, categorical evaluation of the reasoning steps in the unstructured textual format is difficult. One might assume that an ideal structured representation of explanation could also be produced by LLMs via in-context prompting, but it has been demonstrated that LLMs have major struggles in generating structured data even via few-shot prompting~\citep{DBLP:journals/corr/abs-2305-12392}. We highlight this further in our experiments. 

%This necessitates more attention on the generation of structured explanations from large language models. 

% \footnote{We show few-shot learning results on generating semi-structured explanations in Appendix~\ref{seg_llms}.}.

%A more mature space of research has focused on investigating the semi-structured explanation generation (SEG) via smaller language model. 
For this purpose, \citet{DBLP:conf/emnlp/SahaYBB21} propose to use multiple models for predicting answer, internal nodes, external nodes and relations. %\citet{DBLP:conf/acl/SahaYB22} also propose to leverage contrastive learning to generate the explanation graph. 
%\citet{DBLP:conf/acl/CuiLZS23} incorporate a generative pre-training mechanism over synthetic graphs on the explanation graph generation \ehsan{this sentence is not clear}. 
\citet{DBLP:conf/acl/CuiLZS23} incorporate a generative pre-training mechanism over synthetic graphs by aligning input pairs of text-graph to improve the model's capability in generating semi-structured explanation. Both works train separate models for prediction of response, and generation of explanations. 
It is reasonable to expect that even a moderately-sized language model such as FLAN-T5-XXL~\cite{DBLP:journals/corr/abs-2210-11416} should be capable of producing both answers and the corresponding structured explanations. We investigate this in our work. In parallel, Large LMs at the scale of GPT-4~\citep{DBLP:journals/corr/abs-2303-08774} have shown a great capability in producing both an answer and an unstructured reasoning trace through methods like Chain-of-Thought~(CoT)~\citep{DBLP:conf/nips/KojimaGRMI22,DBLP:conf/nips/Wei0SBIXCLZ22}. One might hope that an ideal structured representation of the reasoning trace could also be comfortably surfaced via in-context prompting of LLMs. But it has been demonstrated that LLMs struggle to generate structured format output~\citep{DBLP:journals/corr/abs-2305-12392}. We empirically verify this struggle in the context of generating structured explanations. 

Our objective is to equip moderately-sized LMs with the ability to not only provide answers but also generate structured explanations. To facilitate this, we first utilise supervised fine-tuning (SFT) as the de-facto solution. We then turn our focus to RLHF\footnote{Throughout this paper, we use RLHF and RL interchangeably. 
% RLHF is commonly used in the literature for referring to alignment approaches that deploy RL. 
Noting that our framework does not involve human feedback alignment, but leverages the same framework to create a better alignment between LM's predictive behaviour and ground-truth.} as a mechanism to further align the explanations with ground-truth on top of SFT. We design a reward engineering method in RL and explore multiple reward aggregations that leverage both reward modelling and reward metrics. Our proposed method, implemented on the backbone of a FLAN-T5~\citep{DBLP:journals/corr/abs-2210-11416}, achieves new state-of-the-art results on two benchmarks, ExplaGraph~\citep{DBLP:conf/emnlp/SahaYBB21} and COPA-SSE~\citep{DBLP:conf/lrec/BrassardHKI22}. As a byproduct, our empirical comparison also highlight the limitations of LLMs like GPT-4 and GPT-3.5-instruct to succeed at structured explanation generation (SEG). Furthermore, we delve into a discussion on RL for SEG and highlight what reward metrics work better, and spotlight the challenges (i.e., reward hacking) of balancing the dynamic of policy optimization.

We hope the findings of our work to shed light on both challenges and potentials of RL in SEG as well as the broader space of graph generation.

\section{Semi-structured Explanation}
%
%\subsection{Task Definition}
%COMMENT: in EACL, we need to first define structured explanation formally. Also need to review existing structured explanation dataset/tasks in NLP. See section 2 of explagraph. 
%Explanation generation is related to tasks which require some internal reasoning steps. 
Structured explanation refers to a specific form of explanation that highlights the underlying decision-making processes of the model via an explicit representation of relationships between different reasoning factors. In this section, we briefly review different forms of explanations and introduce the semi-structured explanation tasks of our interest.

\subsection{Related Work}

Explanation in Explainable NLP literature~\citep{DBLP:conf/nips/WiegreffeM21} can be categorised into three major types: (I) \textit{Highlight Explanations} are subsets of the input elements which explain a prediction. For textual NLP tasks, the elements are usually words, phrases or sentences. The representative highlight explanations datasets are  WikiQA~\citep{DBLP:conf/emnlp/YangYM15}, HotpotQA~\citep{DBLP:conf/emnlp/Yang0ZBCSM18}, CoQA~\citep{DBLP:journals/tacl/ReddyCM19}, BoolQ~\citep{DBLP:conf/acl/DeYoungJRLXSW20}, which have different granularities from words to sentences. (II) \textit{Free-text Explanations} are texts in natural language that are not constrained to the input elements, hence more expressive and readable. It is a popular explanation type for both textual and visual-textual tasks with benchmarks like VQA-E~\citep{DBLP:conf/eccv/LiTJCL18}, e-SNLI~\citep{DBLP:conf/nips/CamburuRLB18}, WinoWhy~\citep{DBLP:conf/acl/ZhangZS20}, ECQA~\citep{DBLP:conf/acl/AggarwalMAKSG20}. (III)~\textit{Semi-structured Explanations} are a specific format of explanations which are written in natural language but not entirely free-form. Semi-structured explanations have aroused the public attention in recent years because they combine the properties of both highlight and free-text explanations. Semi-structured explanations do not have one unified definition, but represent explanations in a (semi-)structured format. Benchmarks like WordTree~\citep{DBLP:conf/lrec/JansenWMM18, DBLP:conf/lrec/XieTMWMJ20}, eQASC~\citep{DBLP:conf/emnlp/JhamtaniC20}, ExplaGraph~\citep{DBLP:conf/emnlp/SahaYBB21}, COPA-SSE~\citep{DBLP:conf/lrec/BrassardHKI22} fall under this category. 

\subsection{Tasks}
Since WordTree is based on lexically overlapping sentences and eQASC is based on natural language reasoning chain, neither of them have a unified form of semi-structured explanations. In this work, we focus on two semi-structured explanation tasks: ExplaGraph~\citep{DBLP:conf/emnlp/SahaYBB21} and COPA-SSE~\citep{DBLP:conf/lrec/BrassardHKI22}. Both of them are question-answering tasks and the explanations contain concepts and relations in triple format, which are clear to understand and easy to evaluate. We provide a brief overview of them in what follows and an example of each task in Table~\ref{input_demo}. 
%\ehsan{Need to add 1-2 sentences here stating why we are not using WordTree or eQASC.}

\paragraph{ExplaGraph}
Given a belief and an argument, the task requires a model to predict  whether a certain \emph{argument} supports or counters a \emph{belief}. Each instance in the data is also accompanied by a commonsense explanation graph which reveals an internal reasoning process involved in inferring the predicted stance. The explanation graph is a connected directed acyclic graph (DAG), in which the nodes are concepts (short English phrase) and relations are chosen based on ConceptNet~\citep{Liu2004ConceptNetA}. Concepts are either internal (part of the belief or the argument) or external (part of neither but necessary for the explanation). Semantically, the explanation graphs are commonsense-augmented structured arguments that explicitly support or counter the belief.

\paragraph{COPA-SSE}
Given a premise and a question, the task of COPA-SSE is to select from two options the one that more plausibly has a causal relation with the premise, and generate a corresponding semi-structured commonsense explanation. The semi-structured explanation is created by crowd workers, which contains multiple triples in [head, relation, tail] format. The nodes are concepts and relation are from ConceptNet as well. Different from ExplaGraph, the semi-structured explanation in COPA-SSE is not necessary to be a DAG.

The difficulty of these two tasks is that first the model needs to correctly understand the question and answer it, then generate a reasonable and semantically correct semi-structured explanation. The answers are in a form of an unstructured natural language, while the explanations are of structured format. Tasking a model to generate both modalities, as we will show in the experiment section, imposes a major challenge. In this work, we mainly focus on improving the quality of semi-structured explanations.

%The semantic expression structure has a significant discrepancy between semi-structured data and natural language texts. In addition to the conventional supervised fine-tuning, we propose to use reinforcement learning algorithm with our designed reward engineering method to further enhance the capability of the model in generating semi-structured explanation.  

\section{Reward Engineering for SEG}
%\subsection{Reinforcement Learning}
Motivated by the success of reinforcement learning from human feedback (RLHF)~\citep{DBLP:conf/nips/Ouyang0JAWMZASR22,DBLP:journals/corr/abs-2305-14387,DBLP:journals/corr/abs-2307-09288} in LLMs, we propose to use RL for semi-structured explanation generation task. To achieve this, we design a reward engineering method by incorporating different sources of reward. The RLHF typically begins with a pre-trained LLM fine-tuned with supervised learning on a downstream task, namely the SFT model. The process has two phases: the reward modelling phase and the RL fine-tuning phase. Our reward engineering is designed to improve the reward modelling phase. The objective of RL fine-tuning is to optimize the policy model against a reward model. In our work, we use proximal policy optimization (PPO)~\citep{DBLP:journals/corr/SchulmanWDRK17}.

\subsection{Reward Model}
\label{reward_model}
In the reward modelling phase, given the input and a generated output, the reward model, $R_\phi$, generates a single scalar representing its overall quality. To train a reward model, first we need to collect the paired preference data. In this work, we generate the paired data using the SFT model, which is fine-tuned on the semi-structured explanation task. The SFT model generates the outputs from the training data, then we pair the generated output with its corresponding reference. To improve the quality of the paired preference data, we filter out the pairs where the generated output is the same as the reference. In each pair, the reference is regarded as the preferred data. The filtered paired preference data is then used to train the reward model.
\begin{table}[t]
\centering \footnotesize
\begin{tabular}{p{0.45\textwidth}}
\toprule
\Centering\textbf{ExplaGraph} \\
\hline 
\textbf{Input:} \\
Predict the stance and generate an explanation graph given the belief and argument. \\
Belief: \\
People around the world are able to connect thanks to social media.\\
Argument: \\
Before social media existed there was no quick and easy way to connect with others globally.\\
\hline 
\textbf{Output:} \\
support (social media; causes; connection)(connection; used for; people)(people; at location; globally)(connection; made of; fast connection) \\        \specialrule{2.5pt}{1pt}{1pt}
\Centering\textbf{COPA-SSE} \\
\hline 
\textbf{Input:} \\
Given the premise, choose from a or b and generate an explanation graph. \\
Premise: \\
The man woke up late. What happened as a RESULT?\\
a: He missed an appointment with the dentist. \\
b: He made an appointment with the dentist.\\\bottomrule
\textbf{Output:} \\
a [[The man, HasProperty, sleepy], [Sleepiness, Causes, oversleeping], [oversleeping, Causes, missing events], [a dentist appointment, HasProperty, an event]] \\
\hline 
\end{tabular}
\caption{An example of the input-output for each task. The  explanations are presented as a set of triples of [head, relation, tail]. These triples form: a connected graph in the case of ExplaGraph, or a semi-structured set in the case of COPA-SSE.}
\label{input_demo}
\vspace{-2.5mm}
\end{table}
\subsection{Reward Metric}
\label{rm}
In addition to collecting the reward from the reward model, we propose to collect another reward from evaluation metrics. This metric reward can explicitly reflect the quality of the generated output which is naturally complementary to the reward from the reward model. Since the semi-structured explanation is represented in format of a set of triples (i.e., [head, relation, tail]), following the previous work~\citep{DBLP:conf/emnlp/SahaYBB21}, we consider each triple as a sentence and use the existing text matching metrics to calculate the graph matching score. Specifically, BLEU~\citep{Papineni2002BleuAM}, ROUGE~\citep{Lin2004ROUGEAP} and BERTScore~\citep{Zhang2019BERTScoreET} are extended to Graph-BLEU, Graph-ROUGE and Graph-BERTScore. Graph Edit Distance~(GED)~\citep{DBLP:conf/icpram/Abu-AishehRRM15} takes into account the graph structure of the explanation. 

\subsection{Reward Aggregation}
The reward model $R_\phi$ takes input prompt $x$ and generated output $y$, and generates a single scalar $R_\phi(x, y)$. For the metric reward, given the generated output $y$ and the reference $y'$, the evaluation metric $R_m$ is used to calculate a metric score as the reward $R_m(y, y')$. To aggregate two rewards, an important premise is that the order of magnitude of two rewards should not have too much difference (e.g., 0.01 vs 100), otherwise the effect of one reward could be washed away. To regulate this, we explore various aggregation configurations for the final reward $R(x, y, y')$,
%. One aggregation method is to add the two rewards together directly shown in Equation~\ref{eq1} and the other method is to add weights to different rewards when adding them together shown in Equation~\ref{eq2}.
%\begin{equation}
%R(x, y, y')=R_\phi(x, y)+R_m(y, y')
%\label{eq1}
%\end{equation}
\begin{equation}
R(x, y, y')=\alpha R_\phi(x, y)+(1-\alpha)R_m(y, y')
\label{eq2}
\end{equation}
where $\alpha$ is a coefficient to control the weights of different rewards. In RL phase, we use the total reward to provide feedback to the language model. In particular, we formulate the following optimization problem,
\begin{multline}
\max _{\pi_\theta} \mathbb{E}_{x \sim \mathcal{D}, y \sim \pi_\theta(y \mid x)}\left[R(x, y, y')\right] \\
-\beta \mathbb{D}_{\mathrm{KL}}\left[\pi_\theta(y \mid x) \| \pi_{\mathrm{ref}}(y \mid x)\right]
\end{multline}

\noindent where $\beta$ is the KL coefficient controlling deviation from the base reference policy $\pi_{\mathrm{ref}}$ (the initial SFT model). In practice, the language model policy $\pi_\theta$ is also initialised to the initial SFT model.

\section{Experiment}

\subsection{Datasets and Evaluation Metrics}

ExplaGraph~\citep{DBLP:conf/emnlp/SahaYBB21} contains 2,368/398/400 samples as training/dev/test set. Since the labels of the test set are not public, we provide the evaluation results on dev set.\footnote{We have submitted our prediction test set to evaluate and we will update the test evaluation result once we receive it.} As shown in Table~\ref{input_demo}, for ExplaGraph, the instruction we use is "\textit{Predict the stance and generate an explanation graph given the belief and argument.}" We concatenate the instruction with the belief and argument as input, and the output is a stance concatenated with a linearised explanation graph. We use the same evaluation metrics provided in the ExplaGraph~\citep{DBLP:conf/emnlp/SahaYBB21}: Stance Accuracy~(SA), Structural Correctness Accuracy of Graphs~(StCA), Semantic Correctness Accuracy of Graphs~(SeCA), Graph-BertScore~(G-BS), Graph Edit Distance~(GED), Edge Accuracy (EA). 

COPA-SSE~\citep{DBLP:conf/lrec/BrassardHKI22} contains 1,000/500 samples as training/test set. Since each instance in COPA-SSE contains multiple human-rating semi-structured explanations, we only use the one with the highest rating score as the reference. For COPA-SSE, the instruction we use is "\textit{Given the premise, choose from a or b and generate an explanation graph.}" This instruction is concatenated with the premise and two options as input. The output is the answer along with a semi-structured explanation. 
%We show an example of each task in the Appendix~\ref{input_demo}. 
For evaluation, we use Answer Accuracy~(AA), Triple Match F1 Score~(T-F1), Graph Match F1 Score~(G-F1), Graph-BertScore~(G-BS), Graph Edit Distance~(GED). 

The detailed descriptions of all evaluation metrics are provided in Appendix~\ref{metric_expla}.

\subsection{Models}
\paragraph{LLM Baselines.} To probe the capability of LLMs on generating semi-structured explanations, we conducted experiments on two advanced LLMs, ChatGPT (\textit{gpt-3.5-turbo-instruct}) and GPT-4 (\textit{gpt-4}). We performed 2-shot and 6-shot in-context learning. In addition to the standard prompting we also prompted the models by providing the list of relation types (giving LLM a higher chance of extracting the right relations) in ExplaGraph dataset. The full prompts used for these two tasks are shown in Appendix~\ref{seg_llms}.

\paragraph{SFT.} For supervised fine-tuning (SFT), we conduct experiments on decoder-only architecture models, LLAMA2~\citep{DBLP:journals/corr/abs-2307-09288}, and encoder-decoder architecture models,  FLAN-T5~\citep{DBLP:journals/corr/abs-2210-11416}. For LLAMA2, we use LLaMA2-7B and LLaMA2-13B, and for FLAN-T5, we use FLAN-T5-Large, FLAN-T5-XL and FLAN-T5-XXL. We perform instruction-tuning on the models using LoRA~\citep{DBLP:conf/iclr/HuSWALWWC22}, which is a parameter-efficient fine-tuning method. 
\begin{table}[t]\footnotesize
\setlength{\tabcolsep}{3.2pt}
\centering
\resizebox{\columnwidth}{!}{%
\begin{tabular}{llcccccc}
\toprule
      && Answer                & \multicolumn{5}{c}{Explanation}  \\\cmidrule(lr){3-3}\cmidrule(lr){4-8}
&\multicolumn{1}{l}{}& \textbf{SA↑}   & \textbf{StCA↑} & \textbf{SeCA↑} & \textbf{G-BS↑} & \textbf{GED↓} & \textbf{EA↑}   \\ 
\multicolumn{2}{l}{RE-SP~\citep{DBLP:conf/emnlp/SahaYBB21}}                                                                      & 72.30          & \textbf{62.30} & 18.50          & 47.00 & 0.62          & 27.10          \\
\multicolumn{2}{l}{T5-Large~\citep{DBLP:conf/acl/SahaYB22}}                                                                   & 86.20          & 46.50          & 31.60          & 36.80          & 0.66          & 26.80          \\
\multicolumn{2}{l}{T5-Large + CL~\citep{DBLP:conf/acl/SahaYB22}} & 86.20          & 52.70          & 37.90          & 41.70          & 0.62          & 29.80          \\
\multicolumn{2}{l}{BART-Large~\citep{DBLP:conf/acl/CuiLZS23}}                                                                & 88.19          & 36.43          & 26.13          & 28.42          & 0.74          & 20.77          \\
\multicolumn{2}{l}{BART-Large + EG3P~\citep{DBLP:conf/acl/CuiLZS23}}    & 88.19          & 48.99          & 37.43          & 38.73          & 0.65          & 25.03          \\ \specialrule{1.5pt}{1pt}{1pt}
 \parbox[t]{0mm}{\multirow{4}{*}{\rotatebox[origin=c]{90}{FS=2}}} &ChatGPT (gpt-3.5-turbo-instruct)   & 76.63             & 7.79           & 2.76           & 6.23           & 0.95          & 3.90         \\
 &\ \ \ \ + relation & 73.62             & 20.85          & 4.27           & 16.17          & 0.86          & 10.89        \\
 &GPT-4 (gpt-4)      &  \CCR{30}\textbf{95.73}             & 6.53           & 2.01           & 5.16           & 0.95          & 4.63         \\ 
&\ \ \ \ + relation & 94.47             & 19.60          & 6.53           & 15.31          & 0.86          & 12.62        \\ \cdashline{2-8}
%\multicolumn{7}{c}{6-shot LLMs} \\ 
\parbox[t]{0mm}{\multirow{4}{*}{\rotatebox[origin=c]{90}{FS=6}}} &ChatGPT (gpt-3.5-turbo-instruct)     & 78.89             & 11.56          & 3.76           & 9.09           & 0.92          & 5.77         \\
&\ \ \ \ + relation & 79.65             & 21.11          & 4.32           & 16.66          & 0.86          & 11.13        \\
 &GPT-4 (gpt-4)        & 95.48             & 22.11          & 13.07          & 17.55          & 0.84          & 13.83        \\
&\ \ \ \ + relation & 94.97             & \CCR{30}27.89          & \CCR{30}13.81          & \CCR{30}21.45          & \CCR{30}0.81          & \CCR{30}18.48        \\ \specialrule{1.5pt}{1pt}{1pt}
\parbox[t]{0mm}{\multirow{5}{*}{\rotatebox[origin=c]{90}{SFT}}}
&LLaMA2-7B  & 88.69         & 40.95          & 23.87          & 31.05          & 0.71          & 26.68          \\
&LLaMA2-13B    & 89.45          & 43.72          &  26.38      &  33.86       & 0.69          & 27.62          \\ \cdashline{2-8}
&FLAN-T5-Large-780M $\circ$ & 77.64          & 22.11          & 13.07          & 16.41          & 0.85          & 14.03          \\
&FLAN-T5-XL-3B $\diamond$  & 90.45          & 38.19          & 27.63          &  29.39         & 0.73          & 26.42          \\
&FLAN-T5-XXL-11B $\star$ & \CC{30}{91.71}          & \CC{30}{46.98}          & \CC{30}{35.18}          & \CC{30}{36.14}          & \CC{30}{0.66}          & \CC{30}{31.23}          \\ \specialrule{1.5pt}{1pt}{1pt}
\parbox[t]{0mm}{\multirow{12}{*}{\rotatebox[origin=c]{90}{SFT + RL}}}&$\circ$ + RL with only $R_\phi$                                                                              & 77.39          & 22.11          & 13.07          & 18.09          & 0.84          & 15.40          \\
&$\circ$ + RL with only $R_m$                                                                              & 78.39            & 21.36          & 13.57          & 16.33          & 0.84         & 14.40          \\
&$\circ$ + RL with $R_\phi$, $R_m$    w/o weights                                                               & 78.89 & 25.63         & 16.33 & 20.36 & 0.81 & 16.98 \\ 
&$\circ$ + RL with $R_\phi$, $R_m$    with weights                                                               &79.40 & 24.87 & 15.08& 20.12 & 0.82& 17.00\\ \cdashline{2-8} 
&$\diamond$ + RL with only $R_\phi$                                                                              & 90.45         & 49.25         & 36.18          & 38.92         & 0.64          & 34.67          \\
&$\diamond$ + RL with only $R_m$                                                                              & 90.45          & 40.70         & 28.73          & 31.36         & 0.71          & 28.14          \\
&$\diamond$ + RL with $R_\phi$, $R_m$   w/o weights                                                                  & 90.95 & 50.50          & 36.38 & 39.60 & 0.63 & 36.39\\ &$\diamond$ + RL with $R_\phi$, $R_m$    with weights                                                               & 89.45&46.98 &34.67 &37.55  &0.66 & 32.64\\ \cdashline{2-8} 
&$\star$ + RL with only $R_\phi$                                                                              & 91.46          & 57.54          & 44.47          & 44.83          & 0.59          & 39.38          \\
&$\star$ +  RL with only $R_m$                                                                              & 91.96          & 59.55          & 46.73          & 47.28          & 0.57          & 38.61          \\
&$\star$ +  RL with $R_\phi$, $R_m$ w/o weights                                                                    & \CCG{50}{91.96} & \CCG{50}61.81          & \CCG{50}\textbf{48.49} & \CCG{50}\textbf{47.50} & \CCG{50}\textbf{0.56} & \CCG{50}\textbf{44.16} \\
&$\star$ + RL with $R_\phi$, $R_m$ with weights                                                                   & 91.46          & 56.03          & 42.46          & 44.25          & 0.60          & 38.67    \\  \bottomrule
\end{tabular}%
}
\caption{The evaluation results on ExplaGraph dev set. The $\alpha$ used in "with weights" is 0.9. \textbf{Bold} shows the best result for a column, and arrows indicate the direction of improvement, i.e., ↑: higher is better. Colors denote the best within each group of methods.}
\label{explagraph-main-result} 
\vspace{-2.5mm}
\end{table}
\paragraph{RL.} Previous work has shown that the encoder-decoder architecture models generally perform better than decoder-only architectures in transduction tasks that need a deep understanding of the input~\citep{DBLP:journals/corr/abs-2304-04052}. This finding is in line with our results ~\ref{exp_result}. Therefore, we only use FLAN-T5 models as our backbone models for RL. For reward modelling, since it does not need to perform the down-stream tasks directly, we use LLaMA-7B for simplicity. Inspired by the previous work~\citep{DBLP:journals/corr/abs-2307-09288}, we first fine-tune the pre-trained LLaMA-7B on the task data, then the reward model is initialised from the fine-tuned LLaMA-7B model checkpoint. This can help the reward model to better understand the input and improve the performance. The training details are provided in the Appendix~\ref{training_details}.

\paragraph{Other Baselines.} For ExplaGraph, all of these baselines fine-tune a RoBERTa model to predict the stance label by conditioning on the belief and argument. For explanation graph generation, RE-SP~\citep{DBLP:conf/emnlp/SahaYBB21} combines different models to predict the internal nodes, external nodes and relations, respectively. T5-Large~\citep{DBLP:conf/acl/SahaYB22} and BART-Large~\citep{DBLP:conf/acl/CuiLZS23} generate explanation graphs as post-hoc explanations by conditioning on the belief, argument and the predicted stance using T5-Large model and BART-Large model. T5-Large+CL~\citep{DBLP:conf/acl/SahaYB22} further implements contrastive learning methods on T5-Large. BART-Large+EG3P~\citep{DBLP:conf/acl/CuiLZS23} incorporates a generative pre-training mechanism over synthetic graphs on BART-Large to improve the model’s capability for SEG task. For COPA-SSE, since it is a relatively new dataset, there are no public baselines we can compare.

%\subsection{Evaluation Metrics}
%For ExplaGraph evaluation, we use the same evaluation metrics provided in the ExplaGraph~\citep{DBLP:conf/emnlp/SahaYBB21}: Stance Accuracy~(SA), Structural Correctness Accuracy of Graphs~(StCA), Semantic Correctness Accuracy of Graphs~(SeCA), Graph-BertScore~(G-BS), Graph Edit Distance~(GED), Edge Accuracy (EA). 

%For COPA-SSE evaluation, we use Answer Accuracy~(AA), Triple Match F1 Score~(T-F1), Graph Match F1 Score~(G-F1), Graph-BertScore~(G-BS), Graph Edit Distance~(GED). The detailed descriptions of the above metrics are provided in Appendix~\ref{metric_expla}.

\begin{table}[t]
\setlength{\tabcolsep}{3.2pt}
\centering
\resizebox{\columnwidth}{!}{%
\begin{tabular}{llccccc}
\toprule
      && Answer               & \multicolumn{4}{c}{Explanation} \\ \cmidrule(lr){3-3}\cmidrule(lr){4-7}
&\multicolumn{1}{l}{} & \textbf{AA↑}  & \textbf{T-F1↑} & \textbf{G-F1↑} & \textbf{G-BS↑} & \textbf{GED↓}  \\ 
\parbox[t]{1mm}{\multirow{2}{*}{\rotatebox[origin=c]{90}{FS=2}}} &ChatGPT (gpt-3.5-turbo-instruct)  & 94.8              & 0.55           & 0.00           & 43.99          & 45.79         \\
 &GPT-4 (gpt-4)  & \CCR{30}\textbf{100.0}             & 1.29           & 0.00           & 59.97          & 34.89         \\ \cdashline{2-7}
\parbox[t]{1mm}{\multirow{2}{*}{\rotatebox[origin=c]{90}{FS=6}}}  &ChatGPT (gpt-3.5-turbo-instruct) & 93.4              & 0.85           & 0.00           & 47.86          & 45.55         \\
 &GPT-4 (gpt-4)   & 99.8              & \CCR{30}2.19           & 0.00           & \CCR{30}62.41          & \CCR{30}31.36       \\ \specialrule{1.5pt}{1pt}{1pt}
 \parbox[t]{0mm}{\multirow{5}{*}{\rotatebox[origin=c]{90}{SFT}}}
&LLaMA2-7B  & 60.8 & 1.21 & 8.20 & 63.97 & 19.93\\
&LLaMA2-13B & 83.8 & 1.39 & 8.40 & 65.40 & 19.85 \\ \cdashline{2-7}
&FLAN-T5-Large-780M &88.0&0.93&5.91  & 65.67&20.05 \\
&FLAN-T5-XL-3B &95.4&1.73& 8.39 & \CC{30}\textbf{69.25}& 20.00 \\
&FLAN-T5-XXL-11B $\star$    & \CC{30}97.4          & \CC{30}1.87           & \CC{30}8.42          & 67.20          & \CC{30}19.77          \\\specialrule{1.5pt}{1pt}{1pt}
\parbox[t]{0mm}{\multirow{4}{*}{\rotatebox[origin=c]{90}{SFT + RL}}}
&$\star$ + RL with only $R_\phi$        & \CCG{50}98.0          & 2.01           & 11.71         & 67.93         & 18.65          \\
&$\star$ + RL with only $R_m$         & 97.2          & 1.93           & 10.85          & 67.50          & 19.02          \\
&$\star$ + RL with $R_\phi$, $R_m$ w/o weights           & 97.8 & \CCG{50}\textbf{2.33}  &\CCG{50}\textbf{12.47} & \CCG{50}68.80 & \CCG{50}\textbf{17.49} \\ 
&$\star$ + RL with $R_\phi$, $R_m$ with weights          & 97.2 & 2.05 & 10.87 & 67.68 & 18.75 \\ \bottomrule  

\end{tabular}%
}
\caption{The evaluation results on COPA-SSE test set. The weight factor $\alpha$ used in last setting is 0.5. \textbf{Bold} shows the best result for a column, and arrows indicate the direction of improvement, i.e., ↑: higher is better. Colors denote the best within each group of methods.}
\label{copa-sse-main-result}
\vspace{-2.5mm}
\end{table}
\subsection{Results}
\label{exp_result}
\paragraph{ExplaGraph.} We demonstrate the evaluation results on ExplaGraph in Table~\ref{explagraph-main-result}, comparing with other baseline methods. For SFT results, FLAN-T5-XXL performs better than LLaMA2-13B. As the model size increases, the performance also improves accordingly. Even only doing SFT on FLAN-T5-XXL can achieve higher SA and EA than all five baseline methods. For the RL results, when we only use single reward $R_m$ or $R_\phi$ in RL, the performance is improved. The improvements are much more remarkable in FLAN-T5-XL and FLAN-T5-XXL. The metric reward we use is G-BERTScore (see \S\ref{ab:metric} for ablation on the metrics) and the KL coefficient $\beta$ is 0.3 (see \S\ref{ab:coef} for ablation on coefficients) for RL, which are the best setting based on our experiments.

Using single metric reward $R_m$ is more effective than using the reward model $R_\phi$ on FLAN-T5-XXL. The aggregation of $R_\phi$ and $R_m$ without using weights performs better than with weights on all three FLAN-T5 models. FLAN-T5-XXL achieves the best results outperforming the baselines on four metrics by a large margin. Since we did not add any constraints on the structure of predicted graph comparing with the RE-SP~\citep{DBLP:conf/emnlp/SahaYBB21} baseline method which explicitly enforces graph structure constraints (i.e., connectivity and acyclicity), this could explain why StCA is not the highest for our method. The aggregation of two rewards using weight performs even worse than using single reward. We speculate that using weight decreases the effect of two rewards, thus leading to an undesired influence to the RL. 

\paragraph{COPA-SSE.} The evaluation results on COPA-SSE is shown in Table~\ref{copa-sse-main-result}. Using RL can steadily improve the performance of the SFT model, especially when conducting rewards aggregation without using weights. This is consistent with the result shown on ExplaGraph dataset.

%To probe the capability of LLMs on generating semi-structured explanation, we conducted experiments on two advanced LLMs, ChatGPT and GPT-4. The specific model names used in OpenAI API is \textit{gpt-3.5-turbo-instruct} and \textit{gpt-4}, respectively. We performed 2-shot learning and 6-shot learning on both ExplaGraph dev set and COPA-SSE test set. Additionally, since there are only 28 relation types in ExplaGraph dataset, we can also specify the all relations when prompting the LLMs. The full prompts used for these two tasks are shown in Appendix~\ref{seg_llms}. 

\paragraph{Performance of LLMs.} The GPT-4 performs far better than ChatGPT both in answer prediction and explanation generation, which reveals GPT-4 has a stronger reasoning ability than ChatGPT. Including the relation information (denoted as +relation) can greatly improve the performance in both models. Surprisingly, the stance accuracy on GPT-4 using few-shot learning has surpassed the SFT models. However, even using 6-shot learning on LLMs, the performance on SEG is still far behind the SFT models. For COPA-SSE task, GPT-4 even achieves 100\% accuracy on answer prediction using 2-shot learning. However, when using 6-shot learning, the answer accuracy drops a little bit on both GPT-4 and ChatGPT models, although the quality of explanation increases. We speculate that adding more demonstrations introduces some extra information which may affect the model's judgement on answer prediction. G-F1 score is 0 on all settings, which means none of the generated semi-structured explanation matches exactly to the gold reference. This indicates the challenge of generating semi-structured explanation on LLMs and provides a direction for future research.
\begin{figure}[t]
\centering
\begin{subfigure}[b]{0.25\textwidth}
\includegraphics[scale=0.4,trim={2.8cm 1.3cm 1.5cm 0.5cm},clip]{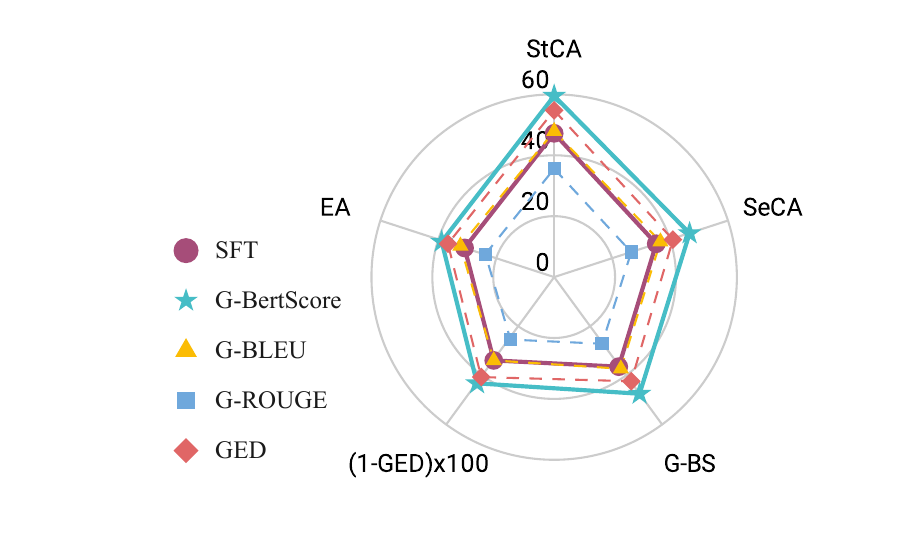}
\end{subfigure}%
\begin{subfigure}[b]{0.25\textwidth}
    ~~~~~~\includegraphics[scale=0.4,trim={6.25cm 1.3cm 1.5cm 0.5cm},clip]{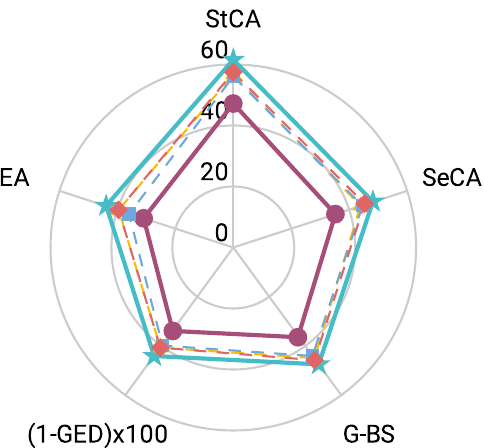}
\end{subfigure}
\caption{Comparison, on ExplaGraph, of SFT and various RL configurations to calculate $R_m$. The KL Coefficient $\beta$ is 0.3 for all experiments. (left) RL using only reward metric, (right) RL using both reward model and metric without any weights. }
\label{figure:diff_rm}
\vspace{-4mm}
\end{figure}

\subsection{Effect of Different Metrics in $R_m$}\label{ab:metric}
In Section~\ref{rm}, we introduced four metrics Graph-BLEU, Graph-ROUGE and Graph-BERTScore, and Graph Edit Distance which could be used to calculate $R_m$. To probe the effect of these metrics, we conduct probing experiments on ExplaGraph. The results are shown in Figure~\ref{figure:diff_rm} (Full results provided in Table~\ref{diff_rm} of Appendix). Graph-BERTScore performs best among all metrics. We speculate this is because the BLEU and ROUGE are calculated using overlapping n-grams. Essentially for the graph-structured data containing multiple triples, the calculation of n-grams becomes less meaningful. However, Graph-BERTScore is a semantic evaluation metric which is still useful in graph-structured data, thus leading to better performance in $R_m$. Interestingly, GED - which considers the structure of the explanation - as a reward metric is not as effective as Graph-BERTScore. This echos the challenge of identifying sources of feedback for RLHF that align well with the underlying task specification~\cite{DBLP:journals/corr/abs-2307-15217}.

%A satisfying discussion of this is beyond the focus of our work and we leave it to future work.

\begin{figure}[t]
\centering
\begin{subfigure}[b]{\columnwidth}
\centering
\includegraphics[scale=0.5,trim={7.8cm 7.7cm 7cm 7.1cm},clip]{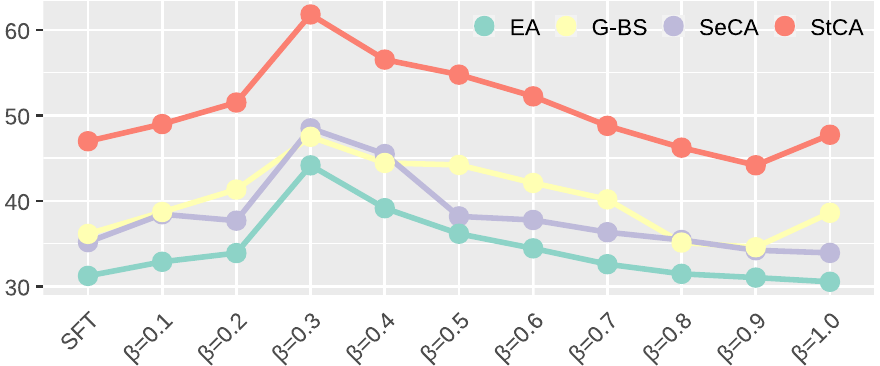}
\caption{Different $\beta$ values.}
    \label{fig:beta}
\end{subfigure}

\begin{subfigure}[b]{\columnwidth}
\centering
    \includegraphics[scale=0.5,trim={7.8cm 7.7cm 7cm 7.1cm},clip]{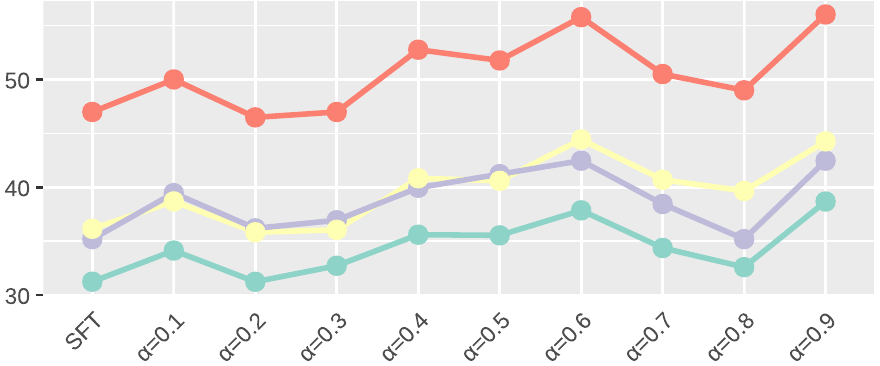}
    \caption{Different $\alpha$ values.}
    \label{fig:alpha}
\end{subfigure}
\caption{FLAN-T5-XXL - SFT in comparison (on ExplaGraph dev set) with SFT+RL under (a) different values of KL Coefficient $\beta$ (we use the aggregation method without weights), and (b) different values of weight factor $\alpha$ (fixing $\beta=0.3$).}
\vspace{-2.5mm}
\end{figure}

\subsection{Effect of $\beta$ and $\alpha$ Coefficients}\label{ab:coef}
\textbf{KL Coefficient $\beta$} is a significant parameter controlling the deviation from the SFT model. To investigate the effect of $\beta$, we conduct experiments on ExplaGraph dataset using different values of $\beta$ (from 0.1 to 1.0). The results are demonstrated in Figure~\ref{fig:beta} (See Table~\ref{kl_beta} in Appendix~\ref{results_details} for full results). As the $\beta$ increases from 0.1, the performance becomes better until $\beta$ is over 0.3. From 0.3 to 1.0, the performance goes down gradually, although they achieve the highest SA. In general, setting $\beta$ as 0.3 leads to the best performance in both ExplaGraph and COPA-SSE tasks. When $\beta$ is small (e.g., 0.1) the new model deviates far from the old model. In this case, although there is a slight improvement, the model  may also learn some undesired pattern to achieve higher rewards (i.e., reward hacking). As the $\beta$ increases, it forces the new model to remain close to the old model, leading a steady improvement. When $\beta$ is close to 1.0, the performance is almost identical to SFT.
%\ehsan{Any speculation here?}

%\subsection{Effect of Weight Factor $\alpha$}
\noindent\textbf{Weight factor $\alpha$} in our reward aggregation method is used to control the importance of different rewards. Although using the reward aggregation method without weights (i.e., removing $\alpha$ and $1-\alpha$)  performs better, here we investigate the effect of $\alpha$ (from 0.1 to 1.0). The results are shown in Figure~\ref{fig:alpha} (See Table~\ref{alpha} in Appendix~\ref{results_details} for full results). From the results, there is no explicit pattern, but in general, larger values of $\alpha$ result in better performance. This means in reward aggregation, the reward from reward model $R_\phi$ is more significant than metric reward $R_m$. A dynamic adaptation of $\alpha$ depending on instances is an interesting direction to investigate in future.
\begin{table}[t]
\scriptsize
\centering
\resizebox{\columnwidth}{!}{%
\begin{tabular}{lcccc}
\toprule
&Rank 1st&Rank 2nd&Rank 3rd& Avg. Rank\\
\textbf{Gold}& 87 & 38 & 75 & 1.94   \\
\textbf{SFT}& 46 & 93& 61& 2.08  \\
\textbf{SFT+RL}& 67  & 69  & 64 & 1.98\\
%\multicolumn{1}{c}{} & \textbf{Gold} & \textbf{SFT} & \textbf{SFT} \\ \hline
%Rank 1st    & 87                    & 46                  & 67                       \\
%Rank 2nd    & 38                    & 93                  & 69                       \\
%Rank 3rd    & 75                    & 61                  & 64                       \\ \hline
%Avg. Rank    & 1.94                  & 2.08                & 1.98                    \\ 
%\hline
%Cohen's Kappa & & 0.45 &
\bottomrule
\end{tabular}
}
\caption{Human evaluation results on 100 ExplaGraph samples by 2 assessors (200 evaluations in total).}
\label{human_eval_result}
\vspace{-4mm}
\end{table}
\section{Analysis}

\subsection{Human Evaluation}
To further evaluate the quality of the generated output from SFT and SFT+RL models, we conduct a human evaluation on 100 randomly sampled instances from ExplaGraph which have correct stance predictions. For each instance, given a belief, an argument and its corresponding stance, we provide assessors with three explanation graphs: Gold reference, SFT, and SFT+RL output. For the evaluation process we recruited two annotators (with at least Master's degree in NLP). Assessors were instructed to rank the three explanation graphs without disclosing their sources, based on the quality of each graph. The human evaluation (total of 200 evaluations) results are demonstrated in Table~\ref{human_eval_result}. As expected, Gold reference ranks first most of the time, followed by SFT+RL output, then SFT output. Based on the average ranking, the SFT+RL output has a higher ranking than the SFT output and a small gap with the gold reference. This indicates that using RL can improve the quality of the generated semi-structured explanation graphs. To our surprise, gold reference has the highest third ranking. Since the ground-truth is created by human annotators, it is inevitably influenced by subjectivity\footnote{Cohen's $\kappa$ of our human evaluation result is 0.18±0.15 with confidence 95\% indicating a slight agreement, which also underscores the subjectivity of the explanation task.}. This necessitates the human evaluation in addition to the automatic evaluation. 

% We observe this also via Cohen's Kappa $\kappa=0.45$ which underscores moderate agreement between the human assessors.

%According to their ranking results on the 100 samples, we counted the number of times each of the three different explanation graphs (Gold, SFT, SFT+RL) ranking first, second and third respectively, and calculated the average ranking of the different explanation graphs. 

\subsection{Qualitative Examples}
In Figure~\ref{fig:seg} we demonstrate two examples from ExplaGraph. In the first example, SFT output fails to generate the relation between "\textit{natural habitats}" and "\textit{natural environments}", while SFT+RL output generate the relation "\textit{PartOf}". This is important for connecting the belief with the argument in the explanation graph. In the second example, SFT+RL output generates a new concept "\textit{cure disease}" which helps to better understand the function of "\textit{stem cell research}". Additionally, it also increases the chances of generating external concepts even we do not explicitly force the model to do so (i.e., predict the internal and external concepts separately). See more examples in Appendix~\ref{app:qualit}.
%In Table~\ref{qual_exam}, we demonstrate two examples from ExplaGraph. In the first example, SFT output fails to generate the concept "\textit{create people}", while the SFT+RL output is much more complete with regard to an explanation graph given the belief and argument. In the second example, even both of the SFT and SFT+RL outputs can correctly generate the first triple "\textit{(austerity programs; capable of; cut funding)}", SFT+RL output contains the concept "\textit{negative effects}", which is similar to the concept "\textit{hurts business}" in the gold. In general, using RL can make the generated explanation graph more detailed and complete than only using SFT. Additionally, it also increases the chances of generating external concepts even we do not explicitly force the model to do so (i.e., predict the internal and external concepts separately). This finding is consistent with the observations made by a recent work~\citep{Singhal2023ALW} which highlights that RLHF tends to generate similar output, but much longer and with more details comparing to SFT.
%% moved to appendix
\begin{table}[t]
\centering \footnotesize
\resizebox{1\columnwidth}{!}{%
\begin{tabular}{p{0.6\textwidth}}
\toprule 
\Centering\textbf{Triple Level Redundancy} \\
\hline 
\textbf{Belief: } \\
Marriage offers numerous benefits.\\
\textbf{Argument: }\\
Marriage is just a piece of paper. \\
\hline 
\textbf{Output:} \\
counter (marriage; is a; piece of paper)(piece of paper; not capable of; numerous benefits)\textcolor{red}{(piece of paper; not capable of; numerous benefits)}\\
\specialrule{2.5pt}{1pt}{1pt}
\Centering\textbf{Concept Level Redundancy} \\
\hline 
\textbf{Belief: } \\
Entrapment helps solve crimes.\\
\textbf{Argument: }\\
Entrapment violates liberties. \\
\hline 
\textbf{Output:} \\
counter (entrapment; capable of; violates liberties)(violates liberties; not capable of; helps solve crimes)\textcolor{red}{(entrapment; synonym of; entrapment)}\\
\bottomrule 
\end{tabular}
}
\caption{Two types of redundancy errors in SFT+RL outputs. Errors are shown in red color text.}
\label{error}
\vspace{-4mm}
\end{table}
\subsection{Error Analysis}\label{sec:errors}
During the human evaluation process, we collected the errors in SFT+RL outputs. Specifically, there are two types of redundancy errors: Triple Level Redundancy and Concept Level Redundancy. We demonstrate an example of each type in Table~\ref{error}. Triple Level Redundancy means the outputs contain repetitive triples. Based on our observation, the repetitive triple is usually the last triple in the generated explanation graph. In the Triple Level Redundancy example in Table~\ref{error}, the triple "\textit{(piece of paper; not capable of; numerous benefits)}" is generated twice. Concept Level Redundancy means the outputs contain repetitive concepts. This type of error is usually associated with a specific relation "\textit{synonym of}". In the Triple Level Redundancy example in Table~\ref{error}, the triple "\textit{(entrapment; synonym of; entrapment)}" contains the repetitive concept "\textit{entrapment}". 
We speculate these undesired behaviours emerge during the policy optimization stage in RL. One general solution for these errors is to enhance robustness and generalization of the reward model (e.g., improve the quality of the preference paired data). In addition, one can also explicitly target redundancy in the RL phase (i.e., via metric design or direct penalty on the reward). It is worth noting that this might not be effective in practice due to the rarity of such patterns during the optimization phase. We leave further exploration of these to future.
%For example, we penalise the reward (e.g., reduce the reward value) of a output if it has the redundancy problem. This can explicitly force the model to avoid generating the repetitive contents.

% \begin{figure}[t]
% \centering
% \begin{subfigure}[b]{0.49\textwidth}
% \includegraphics[scale=0.13]{figures/reward2.png}
% \caption{Mean reward plot during RL training on ExplaGraph using two values of KL Coefficient $\beta$.}
%     \label{fig:reward}
% \end{subfigure}
% \begin{subfigure}[b]{0.49\textwidth}
%     \includegraphics[scale=0.13]{figures/kl2.png}
%     \caption{KL plot during RL training on ExplaGraph using two values of KL Coefficient $\beta$.}
%     \label{fig:kl}
% \end{subfigure}
% \caption{An illustration of the mean reward and the kl during RL training on ExplaGraph: (a) as the training continues, the rewards of both settings increase. While in (b) when $\beta$ is 0.1, the large KL indicates significant deviation from the original SFT model, thus leading to a reward hacking phenomenon.}
% \label{fig:reward_kl}
% \vspace{-2mm}
% \end{figure}
\begin{figure}[t]
\centering
\begin{subfigure}{.24\textwidth}
  \centering
  \includegraphics[height=0.8\columnwidth,width=0.95\columnwidth]{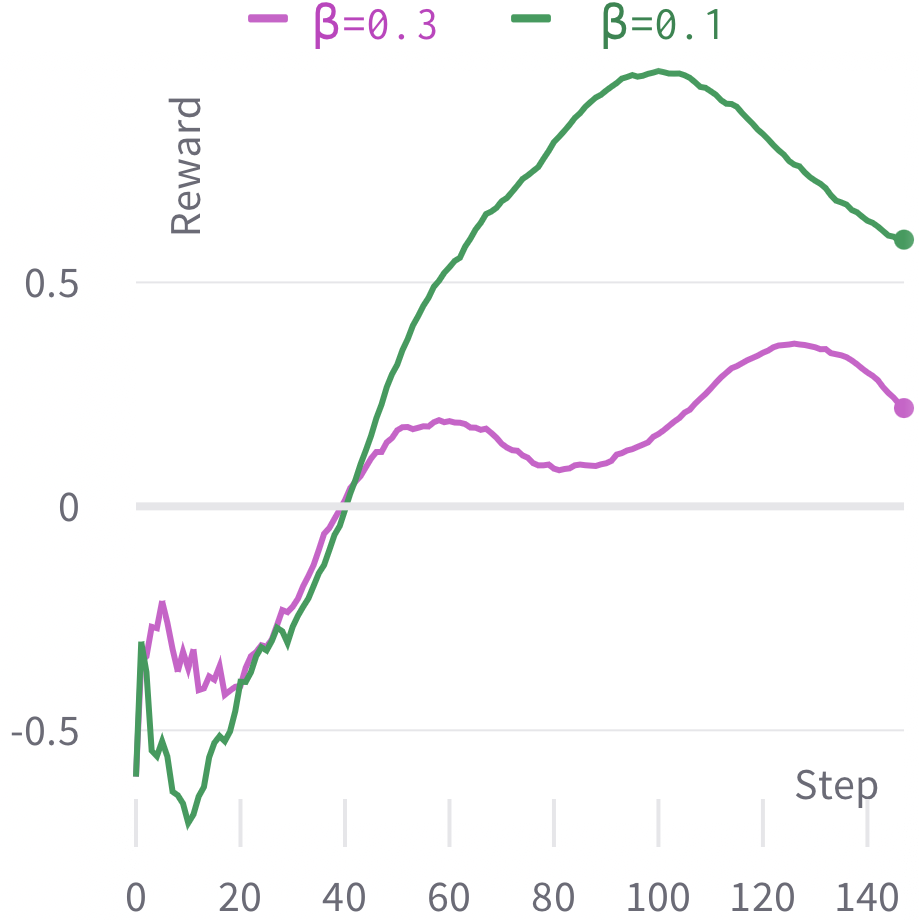}
  \caption{Mean reward plot.}
  \label{fig:reward}
\end{subfigure}%
\begin{subfigure}{.24\textwidth}
  \centering
  \includegraphics[height=0.8\columnwidth,width=0.95\columnwidth]{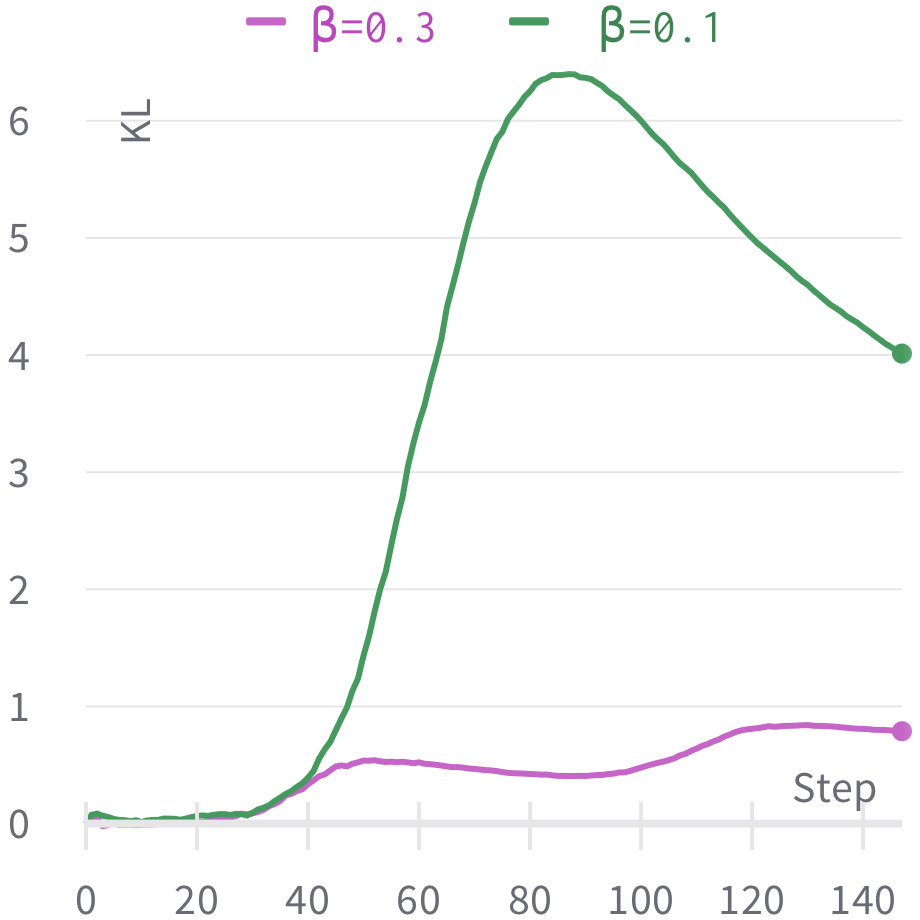}
  \caption{KL plot.}
  \label{fig:kl}
\end{subfigure}
\caption{An illustration of the mean reward and the kl during RL training on ExplaGraph: (a) as the training continues, the rewards of both settings increase. While in (b) when $\beta$ is 0.1, the large KL indicates significant deviation from the original SFT model, thus leading to a reward hacking phenomenon.}
\label{fig:reward_kl}
\vspace{-4mm}
\end{figure}

\subsection{Reward Hacking}
Reward hacking~\citep{DBLP:journals/corr/abs-2209-13085} is a phenomenon where a model achieves high rewards by optimizing a reward function but leading to a low evaluation score on the outputs. Previous work have shown that reward hacking could happen in RLHF training on LLMs~\citep{DBLP:journals/corr/abs-2309-10202, DBLP:journals/corr/abs-2307-09288}. The second term in Eq.~\ref{eq2} is a constraint useful for training stability and mitigating the risk of reward hacking. We demonstrate a mean reward plot and a KL plot in Figure~\ref{fig:reward_kl} to showcase that the RL training with small KL Coefficient $\beta$ (i.e., 0.1) leads to reward hacking. We demonstrates an example showing different outputs from these two settings in Table~\ref{tab:reward_hacking_output}. Under $\beta=0.1$, the model tends to generate longer texts  with unnecessary information. It is worth mentioning that the choice of KL Coefficient depends on different tasks. As discussed earlier (\S\ref{sec:errors}), this stands out as one of the inherent challenges of RLHF application to this task, and choosing a proper KL Coefficient  has a potential in addressing this to some degree. 

Additionally, we observe the average number of triples for SFT and SFT+RLHF on ExplaGraph to be roughly the same (SFT: 3.0$\pm$0.56, SFT+RLHF: 3.1$\pm$0.33). This finding seems to differ from observations in a recent study on text generation~\citep{Singhal2023ALW} which highlights that RLHF tends to generate much longer outputs compared to SFT. We speculate this observation could be an artefact of mild reward hacking, in which a longer sequence could collect further reward via redundancy.
\begin{table}[t]
\centering \footnotesize
\resizebox{1\columnwidth}{!}{%
\begin{tabular}{p{0.6\textwidth}}
\Centering \\ \toprule
\textbf{Belief: } \\
Cosmetic surgery should be banned.\\
\textbf{Argument: }\\
Cosmetic surgery is not worth the risk. \\
\hline 
\textbf{Gold:} \\
support (cosmetic surgery; is a; risky)(risky; used for; human body)(human body; has property; precious)(precious; desires; banned)(banned; used for; risk)\\ \hline
\textbf{SFT+RL ($\beta=0.3)$:} \\
support (cosmetic surgery; has property; dangerous)(dangerous; desires; banned)(cosmetic surgery; has property; not worth the risk)\\ \hline
\textbf{SFT+RL ($\beta=0.1)$} \\ 
support (cosmetic surgery; is a; dangerous)(dangerous; desires; banned)(cosmetic surgery; is a; not worth the risk)(not worth the risk; desires; banned)\textcolor{red}{(cosmetic surgery; synonym of; plastic surgery)(plastic surgery; synonym of; cosmetic surgery)}\\
\bottomrule
\end{tabular}
}
\caption{An example from ExplaGraph dev set to show the output from the model which encounters reward hacking problem (SFT+RL $\beta=0.1$).}
\label{tab:reward_hacking_output}
\vspace{-3mm}
\end{table}
\section{Conclusion}
In this work, we focused on the semi-structured explanation generation task and proposed to train a single model with SFT+RL to generate both  answers and structured explanations. We highlighted the inadequacy of SFT in performing this complex task, and proposed a carefully designed reward engineering method in RL to better address this problem. We investigated different reward aggregation methods and conduct extensive experiments under different settings to better highlight the dynamic of the RL objective function and reward choices. Our method achieves the new SoTA results on two SEG benchmarks, ExplaGraph and COPA-SSE. We provide detailed analysis from different perspectives and hope these empirical findings will be beneficial for the future research on investigating RL in SEG.

\section*{Limitations}
In this work, we only focused on the online alignment method (i.e., using PPO in RL), while there are other offline alignment approaches to align language models with preference data, like DPO~\citep{DBLP:journals/corr/abs-2305-18290}, PRO~\citep{DBLP:journals/corr/abs-2306-17492}, RRHF~\citep{DBLP:journals/corr/abs-2304-05302}. It is also worth investigating the performance of these methods on SEG tasks. 

%Another limitation is that we did not use decoder-only model (i.e., LLaMA2) as the backbone model for RL, so it remains unknown that how much RL can improve from decoder-only model on SEG tasks.

\section*{Ethics Statement}
Our work uses the existing open-source pre-trained models, as such it could inherit the same ethical concerns which has been widely discussed in the community. We uses the public available datasets which is broadly accepted by the community. 
The created training data from COPA-SSE does not generate any new data, which also do not have the ethical issues.

% \section*{Acknowledgements}
% This document has been adapted by Isabelle Augenstein and Andreas Vlachos from the style files used for earlier ACL, EMNLP and NAACL proceedings, including those for
% EMNLP 2022 by Yue Zhang, Ryan Cotterell and Lea Frermann,
% ACL 2020 by Steven Bethard, Ryan Cotterell and Rui Yan,
% ACL 2019 by Douwe Kiela and Ivan Vuli\'{c},
% NAACL 2019 by Stephanie Lukin and Alla Roskovskaya, 
% ACL 2018 by Shay Cohen, Kevin Gimpel, and Wei Lu, 
% NAACL 2018 by Margaret Mitchell and Stephanie Lukin,
% Bib\TeX{} suggestions for (NA)ACL 2017/2018 from Jason Eisner,
% ACL 2017 by Dan Gildea and Min-Yen Kan, NAACL 2017 by Margaret Mitchell, 
% ACL 2012 by Maggie Li and Michael White, 
% ACL 2010 by Jing-Shin Chang and Philipp Koehn, 
% ACL 2008 by Johanna D. Moore, Simone Teufel, James Allan, and Sadaoki Furui, 
% ACL 2005 by Hwee Tou Ng and Kemal Oflazer, 
% ACL 2002 by Eugene Charniak and Dekang Lin, 
% and earlier ACL and EACL formats written by several people, including
% John Chen, Henry S. Thompson and Donald Walker.
% Additional elements were taken from the formatting instructions of the \emph{International Joint Conference on Artificial Intelligence} and the \emph{Conference on Computer Vision and Pattern Recognition}.

% Entries for the entire Anthology, followed by custom entries
\bibliography{anthology,custom}
\bibliographystyle{acl_natbib}

\appendix
\section*{Appendix}
%\section{Task Input Demonstration}
%\label{input_demo}

\section{Evaluation Metrics}
\label{metric_expla}

\paragraph{Stance Accuracy (SA)} measures the stance prediction accuracy which ensures that the explanation graph is consistent with the predicted stance. 
\paragraph{Structural Correctness Accuracy of Graphs (StCA)} requires satisfying all the constraints defined for the task, which include the graph be connected DAG with at least three edges and having at least two exactly matching concepts from the belief and two from the argument.
\paragraph{Semantic Correctness Accuracy of Graphs (SeCA)} requires all edges to be semantically coherent and given the belief, the unambiguously inferred stance from the graph matches the original stance.
\paragraph{Graph-BertScore (G-BS)} considers graphs as a set of edges and solve a matching problem that ﬁnds the best assignment between the edges in the gold graph and those in the predicted graph. Each edge is treated as a sentence and the scoring function between a pair of gold and predicted edges is given by BERTScore. Given the best assignment and the overall matching score, compute precision, recall and report F1 as the G-BERTScore metric.
\paragraph{Graph Edit Distance (GED)} measures the number of edit operations (addition, deletion, and replacement of nodes and edges) for transforming the predicted graph to a graph isomorphic to the gold graph. The cost of each edit operation is chosen to be 1. The GED for each sample is normalized between 0 and 1 by an appropriate normalizing constant (upper bound of GED). Lower GED indicates that the predicted graphs match more closely with the gold graphs.
\paragraph{Edge Accuracy (EA)} computes the macro-average of important edges in the predicted graphs. An edge is defined as important if not having it as part of the graph causes a decrease in the model’s confidence for the target stance.

\paragraph{Answer Accuracy (AA)} calculates the answer prediction accuracy.

\paragraph{Triple Match F1 Score (T-F1)} calculates F1 score based on the precision-recall between the triples in the generated graph and the ground-truth.
\paragraph{Graph Match F1 Score (G-F1)} focuses on the entirety of the graph and evaluates how many graphs are exactly produced the same.

\section{Full Results}
\label{results_details}
Table~\ref{kl_beta} and Table~\ref{alpha} demonstrate the full results of experiments on ExplaGraph using different values of KL Coefficient $\beta$ and weight factor $\alpha$.

\begin{table}[t]
\centering
\resizebox{\columnwidth}{!}{%
\begin{tabular}{lcccccc}
\toprule
      & Answer               & \multicolumn{5}{c}{Explanation}  \\\cmidrule(lr){2-2}\cmidrule(lr){3-7}
\multicolumn{1}{l}{} & \textbf{SA↑}   & \textbf{StCA↑} & \textbf{SeCA↑} & \textbf{G-BS↑} & \textbf{GED↓} & \textbf{EA↑}   \\ \hline
FLAN-T5-XXL - SFT    & 91.71          & 46.98          & 35.18          & 36.14          & 0.66          & 31.23          \\
+ RL, $\beta=0.1$          & 91.46          & 48.99          & 38.44          & 38.70          & 0.65          & 32.88          \\
+ RL, $\beta=0.2$          & 91.71          & 51.51          & 37.69          & 41.33          & 0.64          & 33.90          \\
+ RL, $\beta=0.3$          & 91.96 & \textbf{61.81}          & \textbf{48.49} & \textbf{47.50} & \textbf{0.56} & \textbf{44.16} \\
+ RL, $\beta=0.4$           & \textbf{92.21}          & 56.53          & 45.48          & 44.44          & 0.59          & 39.15          \\
+ RL, $\beta=0.5$          & \textbf{92.21}          & 54.77          & 38.19          & 44.21          & 0.61          & 36.16          \\
+ RL, $\beta=0.6$         & \textbf{92.21}          & 52.23          & 37.77          & 42.10          & 0.63          & 34.45          \\
+ RL, $\beta=0.7$          & \textbf{92.21}          & 48.78          & 36.34          & 40.18          & 0.65          & 32.60          \\
+ RL, $\beta=0.8$           & \textbf{92.21}          & 46.23          & 35.43          & 35.13          & 0.67          & 31.47          \\
+ RL, $\beta=0.9$          & \textbf{92.21}          & 44.17          & 34.23          & 34.58          & 0.67          & 31.03          \\
+ RL, $\beta=1.0$          & \textbf{92.21}          & 47.74          & 33.92          & 38.61          & 0.66          & 30.54         \\ \hline
\end{tabular}%
}
\caption{The full evaluation results on ExplaGraph dev set using different values of KL Coefficient $\beta$. For the reward aggregation in RL, we use the aggregation method without weights.}
\label{kl_beta}
\end{table}

\begin{table}[t]
\centering
\resizebox{\columnwidth}{!}{%
\begin{tabular}{lcccccc}
\toprule
      & Answer               & \multicolumn{5}{c}{Explanation}  \\\cmidrule(lr){2-2}\cmidrule(lr){3-7}
                  & \textbf{SA↑}   & \textbf{StCA↑} & \textbf{SeCA↑} & \textbf{G-BS↑} & \textbf{GED↓} & \textbf{EA↑}   \\ \hline
FLAN-T5-XXL - SFT  & 91.71          & 46.98          & 35.18          & 36.14          & 0.66          & 31.23          \\
+ RL, $\alpha=0.1$         & 91.96          & 50.00          & 39.45          & 38.68          & 0.64          & 34.12          \\
+ RL, $\alpha=0.2$         & 92.46          & 46.48          & 36.18          & 35.82          & 0.67          & 31.22          \\
+ RL, $\alpha=0.3$        & \textbf{92.21} & 46.98          & 36.93          & 36.04          & 0.66          & 32.71          \\
+ RL, $\alpha=0.4$        & 91.71          & 52.76          & 39.95          & 40.83          & 0.62          & 35.59          \\
+ RL, $\alpha=0.5$         & 91.46 & 51.76          & 41.21 & 40.59 & 0.63 & 35.53 \\
+ RL, $\alpha=0.6$         & 91.71          & 55.78          & \textbf{42.46} & \textbf{44.43} & \textbf{0.60} & 37.85          \\
+ RL, $\alpha=0.7$         & 91.46          & 50.50          & 38.44          & 40.69          & 0.64          & 34.36          \\
+ RL, $\alpha=0.8$         & 91.71          & 48.99          & 35.18          & 39.65          & 0.65          & 32.58          \\
+ RL, $\alpha=0.9$         & 91.46          & \textbf{56.03} & \textbf{42.46} & 44.25          & \textbf{0.60} & \textbf{38.67} \\ \hline
\end{tabular}%
}
\caption{The full evaluation results on ExplaGraph dev set using different values of weight factor $\alpha$. The KL Coefficient $\beta$ used is 0.3 for all experiments.}
\label{alpha}
\end{table}

\begin{table}[t]
\centering
\resizebox{\columnwidth}{!}{%
\begin{tabular}{lcccccc}
\toprule
      & Answer               & \multicolumn{5}{c}{Explanation}  \\\cmidrule(lr){2-2}\cmidrule(lr){3-7}
                  & \textbf{SA↑}   & \textbf{StCA↑} & \textbf{SeCA↑} & \textbf{G-BS↑} & \textbf{GED↓} & \textbf{EA↑}   \\ 
FLAN-T5-XXL - SFT & 91.71          & 46.98          & 35.18          & 36.14          & 0.66          & 31.23          \\\cdashline{1-7}
+ RL with only $R_m$~(G-BS)     & 91.96          & 59.55          & 46.73          & 47.28          & 0.57          & 38.61          \\
+ RL with only $R_m$~(G-BL)    & 91.71          & 47.99          & 36.93          & 36.91          & 0.66          & 32.65          \\
+ RL with only $R_m$~(G-RO)   & \textbf{92.46}          & 35.43          & 26.38          & 26.70          & 0.75          & 23.87          \\
+ RL with only $R_m$~(GED)   &    91.96       &     54.77       &      40.95    &     42.52    &   0.59     &    36.52      \\ \cdashline{1-7}
+ RL with $R_\phi$, $R_m$~(G-BS)   & 91.96 & \textbf{61.81} & \textbf{48.49} & \textbf{47.50} & \textbf{0.56} & \textbf{44.16} \\
+ RL with $R_\phi$, $R_m$~(G-BL)  & 91.96          & 57.04          & 44.22          & 45.20          & 0.59          & 39.54          \\
+ RL with $R_\phi$, $R_m$~(G-RO) & 91.96          & 56.03          & 44.47          & 44.30          & 0.60          & 35.99         \\ 
+ RL with $R_\phi$, $R_m$~(GED) & 92.21 &    57.54     &    45.47      &     45.63    &   0.59   &  39.32 \\ \bottomrule
\end{tabular}%
}
\caption{The evaluation results on ExplaGraph dev set under various metrics to calculate $R_m$. We use the aggregation method without weights. The KL Coefficient $\beta$ is 0.3 for all experiments.}
\label{diff_rm}
\vspace{-2.5mm}
\end{table}
\section{Training Details}
\label{training_details}

All models are implemented using Pytorch~\citep{DBLP:conf/nips/PaszkeGMLBCKLGA19} and Transformers~\citep{DBLP:conf/emnlp/WolfDSCDMCRLFDS20}. We use Adam~\citep{DBLP:journals/corr/KingmaB14} and Adafactor optimizer~\citep{DBLP:conf/icml/ShazeerS18}. For the implementation of parameter efficient training method used in FLAN-T5-XXL and LLaMA-7B, we use PEFT~\citep{peft} and 8-bit quantization technique~\citep{DBLP:conf/iclr/DettmersLSZ22}. All training was done using a single A40 GPU with 48GB of RAM. Table \ref{tab:hype1}, Table \ref{tab:hype2} and Table \ref{tab:hype3} show the hyperparameters for SFT Model, Reward Model and RL model, respectively.

\begin{table}[t]
\centering
\scalebox{1.0}{
\begin{tabular}{lc}
\hline
Hyperparameter  & Assignment \\ \hline
Model  & FLAN-T5-XXL      \\
Epoch  & 5     \\ 
Batch Size  & 16     \\
Optimizer  & adamw_torch     \\ 
Learning Rate & $3\times10^{-4}$     \\ 
Warm-up Step & 50     \\ 
Beam Size & 4    \\ 
Lora-r & 4    \\
Lora-alpha & 16    \\
Lora-dropout & 0.05  \\
Lora-modules & [q, v] \\
\hline
\end{tabular}}
\caption{Hyperparameters of SFT Model}
\label{tab:hype1}
\end{table}

\begin{table}[t]
\centering
\scalebox{1.0}{
\begin{tabular}{lc}
\hline
Hyperparameter  & Assignment \\ \hline
Model  & LLAMA-7B      \\
Epoch  & 5     \\ 
Batch Size  & 16     \\
Optimizer  & adamw_torch     \\ 
Learning Rate & $3\times10^{-4}$     \\ 
Warm-up Step & 50     \\ 
Beam Size & 4    \\ 
Lora-r & 8    \\
Lora-alpha & 16    \\
Lora-dropout & 0.05  \\
Lora-modules & [q, v] \\
\hline
\end{tabular}}
\caption{Hyperparameters of Reward Model}
\label{tab:hype2}
\end{table}

\begin{table}[t]
\centering
\scalebox{1.0}{
\begin{tabular}{lc}
\hline
Hyperparameter  & Assignment \\ \hline
Model  & FLAN-T5-XXL      \\
PPO Epoch  & 3     \\ 
Batch Size  & 16     \\
Optimizer  & adafactor     \\ 
Learning Rate & $1.4\times10^{-5}$     \\ 
Warm-up Step & 50     \\ 
Beam Size & 4    \\ 
Lora-r & 8    \\
Lora-alpha & 16    \\
Lora-dropout & 0.05  \\
Lora-modules & [q, v] \\
Target-KL & 2 \\
KL-coef & 0.3 \\
\hline
\end{tabular}}
\caption{Hyperparameters of RL Model}
\label{tab:hype3}
\end{table}

\section{Prompts used for ChatGPT and GPT-4}
\label{seg_llms}
% To probe the capability of LLMs on generating semi-structured explanation, we conducted experiments on two advanced LLMs, ChatGPT and GPT-4. The specific model names used in OpenAI API is \textit{gpt-3.5-turbo-instruct} and \textit{gpt-4}, respectively. We performed 2-shot learning and 6-shot learning on both ExplaGraph dev set and COPA-SSE test set.

For ExplaGraph task, we use the prompt "\textit{Given a belief and an argument, infer the stance (support/counter) and generate the corresponding commonsense explanation graph that explains the inferred stance.}" followed by a few demonstrations. For including relation setting, we use the the prompt "\textit{Given a belief and an argument, infer the stance (support/counter) and generate the corresponding commonsense explanation graph that explains the inferred stance. The available relations in explanation graph are antonym of, synonym of, at location, not at location, capable of, not capable of, causes, not causes, created by, not created by, is a, is not a, desires, not desires, has subevent, not has subevent, part of, not part of, has context, not has context, has property, not has property, made of, not made of, receives action, not receives action, used for, not used for.}" followed by a few demonstrations. 

% The results are demonstrated in Table~\ref{llms_explagraph}. GPT-4 performs far better than ChatGPT both in answer prediction and semi-structured explanation generation, which reveals GPT-4 has a stronger reasoning ability than ChatGPT. Including the relation information can greatly improve the performance in both models. Surprisingly, the stance accuracy on GPT-4 using few-shot learning has surpassed the SFT models. However, even using 6-shot learning on LLMs, the performance on SEG is still far behind the SFT models. 

For COPA-SSE task, we only use the prompt "\textit{Given the premise, choose from a or b and generate an commonsense explanation graph that explains the answer.}" followed by a few demonstrations.

\section{More Qualitative Examples}\label{app:qualit}
In Table~\ref{qual_exam}, we demonstrate two examples from ExplaGraph. In the first example, SFT output fails to generate the concept "\textit{create people}", while the SFT+RL output is much more complete with regard to an explanation graph given the belief and argument. In the second example, even both of the SFT and SFT+RL outputs can correctly generate the first triple "\textit{(austerity programs; capable of; cut funding)}", SFT+RL output contains the concept "\textit{negative effects}", which is similar to the concept "\textit{hurts business}" in the gold. In general, using RL can make the generated explanation graph more detailed and complete than only using SFT. Additionally, it also increases the chances of generating external concepts even we do not explicitly force the model to do so (i.e., predict the internal and external concepts separately). 

\begin{table}[t]
\centering
\resizebox{1\columnwidth}{!}{%
\begin{tabular}{p{0.665\textwidth}}
\toprule
\textbf{Belief: } \\
Human cloning should be allowed, as it would be a great boon for medical advancements.\\
\textbf{Argument: }\\
It is immoral to create people for the sole purpose of curing others. \\
\hline 
\textbf{Gold:} \\
counter (human cloning; used for; create people)(create people; used for; body parts only)(body parts only; has context; immoral)(immoral; not desires; allowed)\\
\textbf{SFT:} \\
counter (human cloning; capable of; immoral)(immoral; not desires; allowed)(immoral; used for; curing others)\\
\textbf{SFT+RL:} \\
counter (human cloning; capable of; immoral)(immoral; not capable of; allowed)(human cloning; capable of; create people)(create people; capable of; curing others)\\
\specialrule{2.5pt}{1pt}{1pt}
\textbf{Belief: } \\
Austerity programs are terrible for the economy.\\
\textbf{Argument: }\\
Austerity programs cut funding. \\
\hline 
\textbf{Gold:} \\
support (austerity programs; capable of; cut funding)(cut funding; capable of; hurts business)(hurts business; causes; terrible)(terrible; has context; for economy)\\
\textbf{SFT:} \\
support (austerity programs; capable of; cut funding)(cut funding; capable of; bad for economy)(bad for economy; synonym of; terrible)\\
\textbf{SFT+RL:} \\
support (austerity programs; capable of; cut funding)(cut funding; capable of; negative effects)(negative effects; capable of; terrible for the economy)\\
\bottomrule
\end{tabular}
}
\caption{Two examples from ExplaGraph dev set to compare the gold explanation graph with the SFT output and SFT+RL output.}
\label{qual_exam}
\vspace{-2.5mm}
\end{table}

\end{document}